\documentclass[]{interact}

\usepackage{subcaption}

\usepackage{graphicx}

\usepackage{hyperref}
\usepackage{tabularx}
\usepackage{makecell}
\usepackage{float}
\usepackage{multicol}
\usepackage{multirow}
\usepackage{soul}
\usepackage{color}
\usepackage{amsmath,amssymb,enumitem,mathtools}
\usepackage{xcolor}
\usepackage{acro}
\usepackage{pdflscape}
\usepackage{natbib}
  
\bibpunct[, ]{(}{)}{;}{a}{}{,}
\renewcommand\bibfont{\fontsize{10}{12}\selectfont}

\theoremstyle{plain}

\theoremstyle{definition}

\theoremstyle{remark}

\DeclareAcronym{insplad}{
  short=InsPLAD,
  long=Inspection Power Line Asset Dataset,
}
\DeclareAcronym{ap}{
  short=AP,
  long=Average Precision,
}
\DeclareAcronym{auroc}{
  short=AUROC,
  long=Area Under the Receiver Operating Characteristic,
}
\DeclareAcronym{uav}{
  short=UAV,
  long=Unmanned Aerial Vehicle,
}
\DeclareAcronym{cv}{
  short=CV,
  long=Computer Vision,
}
\DeclareAcronym{od}{
  short=OD,
  long=Object Detection,
}
\DeclareAcronym{cplid}{
  short=CPLID,
  long=Chinese Power Line Insulator Dataset,
}
\DeclareAcronym{ss}{
  short=SS,
  long=Semantic Segmentation,
}
\DeclareAcronym{is}{
  short=IS,
  long=Instance Segmentation,
}
\DeclareAcronym{ttpla}{
  short=TTPLA,
  long=Transmission Towers and Power Lines Aerial-Image Dataset,
}
\DeclareAcronym{stnplad}{
  short=STN PLAD,
  long=STN Power Line Asset Dataset,
}
\DeclareAcronym{ic}{
  short=IC,
  long=Image Classification,
}
\DeclareAcronym{ad}{
  short=AD,
  long=Anomaly Detection,
}
\DeclareAcronym{yolo}{
  short=YOLO,
  long=You Only Look Once,
}
\DeclareAcronym{rcnn}{
  short=R-CNN,
  long=Region-Based Convolutional Neural Networks,
}
\DeclareAcronym{gan}{
  short=GAN,
  long=Generative Adversarial Networks,
}
\DeclareAcronym{fcn}{
  short=FCN,
  long=Fully Connected Networks,
}
\DeclareAcronym{mlp}{
  short=MLP,
  long=Multilayer Perceptrons,
}
\DeclareAcronym{tood}{
  short=TOOD,
  long=Task-aligned One-stage Object Detection,
}
\DeclareAcronym{vgg}{
  short=VGG,
  long=Very Deep Convolutional Networks,
}
\DeclareAcronym{coco}{
  short=COCO,
  long=Microsoft Common Objects in Context,
}
\DeclareAcronym{iou}{
  short=IoU,
  long=Intersection over Union,
}
\DeclareAcronym{ssd}{
  short=SSD,
  long=Single Shot MultiBox Detector,
}
\DeclareAcronym{ognet}{
  short=OGNet,
  long=Old is Gold Net,
}
\DeclareAcronym{csflow}{
  short=CS-Flow,
  long=Fully Convolutional Cross-Scale-Flows,
}
\DeclareAcronym{aebad}{
  short=AeBAD,
  long=Aero-engine Blade Anomaly Detection Dataset,
}

\begin{document}


\title{InsPLAD: A Dataset and Benchmark for Power Line Asset Inspection in UAV Images}

\author{
\name{André Luiz Buarque Vieira e Silva\textsuperscript{a}\thanks{CONTACT A. L. B. Vieira e Silva. Email: albvs@cin.ufpe.br}, Heitor de Castro Felix\textsuperscript{a}, Franscisco Paulo Magalhães Simões\textsuperscript{a, d}, Veronica Teichrieb\textsuperscript{a}, Michel dos Santos\textsuperscript{b}, Hemir Santiago\textsuperscript{b}, Virginia Sgotti\textsuperscript{b} and Henrique Lott Neto\textsuperscript{c}}
\affil{\textsuperscript{a}Voxar Labs, Centro de Informática, Universidade Federal de Pernambuco, Recife, Brazil; \textsuperscript{b}In Forma Software, Recife, Brazil; \textsuperscript{c}Sistema de Transmissão Nordeste, Recife, Brazil; \textsuperscript{d}Visual Computing Lab, Departamento de Computação, Universidade Federal Rural de Pernambuco, Recife, Brazil}
}

\maketitle

\begin{abstract}
Power line maintenance and inspection are essential to avoid power supply interruptions, reducing its high social and financial impacts yearly. Automating power line visual inspections remains a relevant open problem for the industry due to the lack of public real-world datasets of power line components and their various defects to foster new research. This paper introduces InsPLAD, a Power Line Asset Inspection Dataset and Benchmark containing 10,607 high-resolution Unmanned Aerial Vehicles colour images. The dataset contains seventeen unique power line assets captured from real-world operating power lines. Additionally, five of those assets present six defects: four of which are corrosion, one is a broken component, and one is a bird's nest presence. All assets were labelled according to their condition, whether normal or the defect name found on an image level. We thoroughly evaluate state-of-the-art and popular methods for three image-level computer vision tasks covered by InsPLAD: object detection, through the AP metric; defect classification, through Balanced Accuracy; and anomaly detection, through the AUROC metric. InsPLAD offers various vision challenges from uncontrolled environments, such as multi-scale objects, multi-size class instances, multiple objects per image, intra-class variation, cluttered background, distinct point-of-views, perspective distortion, occlusion, and varied lighting conditions. To the best of our knowledge, InsPLAD is the first large real-world dataset and benchmark for power line asset inspection with multiple components and defects for various computer vision tasks, with a potential impact to improve state-of-the-art methods in the field. It will be publicly available in its integrity on a repository with a thorough description. It can be found at \url{https://github.com/andreluizbvs/InsPLAD/} or \url{https://drive.google.com/drive/folders/1psHiRyl7501YolnCcB8k55rTuAUcR9Ak?usp=sharing}.
\end{abstract}

\begin{keywords}
Power line inspection; Datasets and evaluation; Object detection; Image classification; Anomaly detection; Deep learning
\end{keywords}


\section*{List of Abbreviations}

\textbf{AD} Anomaly Detection  \\
\textbf{AeBAD} Aero-engine Blade Anomaly Detection Dataset  \\
\textbf{AUROC} Area Under the Receiver Operating Characteristic \\
\textbf{TPR} True Positive Rate \\
\textbf{FPR} False Positive Rate \\
\textbf{COCO} Microsoft Common Objects in Context \\
\textbf{CPLID} Chinese Power Line Insulator Dataset \\
\textbf{CV} Computer Vision \\
\textbf{FCN} Fully Connected Networks \\
\textbf{GAN} Generative Adversarial Networks \\
\textbf{IC} Image Classification \\
\textbf{InsPLAD} Inspection Power Line Asset Dataset \\
\textbf{IoU} Intersection over Union \\
\textbf{IS} Instance Segmentation \\
\textbf{MLP} Multilayer Perceptrons \\
\textbf{OD} Object Detection \\
\textbf{SS} Semantic Segmentation \\
\textbf{STN PLAD} STN Power Line Asset Dataset \\
\textbf{TTPLA} Transmission Towers and Power Lines Aerial-Image Dataset \\
\textbf{UAV} Unmanned Aerial Vehicle \\

\section{Introduction}

Nowadays, power line maintenance and inspection are essential for guaranteeing proper functioning and uninterrupted power supply for human activities. Besides that, blackouts may result in severe social impacts and financial problems for the power supply companies  \cite{bruch2011power,li2020state,nguyen2018automatic}. Moreover, many issues can arise in power line components, i.e., power line assets, due to their continuous use, mainly because they are often inserted in open, uncontrolled environments. Nature and human factors such as strong winds, rains and storms, intense sun, sea air, pollution, high vegetation, and birds, in general, can cause rust, cracks, gaps, short circuits, and even the breakdown of the power line components.

Maintaining regular power line inspections is fundamental. However, they are costly, dangerous, and time-consuming \cite{rahmani2013descriptive,hu2017inspection}. Due to the recent advances in computer vision, automating the inspection process is an attractive alternative since the requirement for human inspectors to climb the transmission towers or analyse massive amounts of inspection images would decrease. In short, automating the inspection process can bring safety and reduce financial costs while providing time gains and increasing inspection frequency.

The importance of public datasets in advancing the state-of-the-art of multiple tasks and applications became clear after their introduction. Some examples are \ac{coco} \cite{lin2014coco}, ImageNet \cite{deng2009imagenet} and, for anomaly detection, the MVTec AD \cite{bergmann2019mvtec} and \ac{aebad} \cite{zhang2023industrial}. 

Power line inspection works are usually financed by companies and government agencies which choose not to disclose their datasets \cite{nguyen2018automatic,siddiqui2018robust,zhang2019learning,yang2019insulator,lei2019intelligent}. This may be to maintain some competitive advantage or other confidentiality issues, which, ultimately, hinders the development of this research topic \cite{nguyen2018automatic,liu2020data,xia2022gan}.

Usually, the inspection process of power lines undergoes multiple steps. A straightforward pipeline may be: 1) Detect power line assets in \ac{uav} inspection images and 2) Classify the assets regarding their conditions, e.g., normal or presenting a known defect. However, defects may present themselves in unlimited shapes and sizes, as well as the absence of asset parts. Therefore, it is only reasonable to catalogue common types of defects. A promising way to deal with the less frequent and even unprecedented defects in the industry is to approach this problem as an unsupervised anomaly detection task, in which defects are treated as anomalies \cite{liu2023deep}. 

We propose the \ac{insplad}, a new dataset for inspecting power line assets to address multiple research gaps in the field. The data come from real-world inspections and consists of 10,607 Full HD RGB images of 17 unique power line asset categories captured in the wild by a UAV under multiple environmental conditions, orientations, and distances (28,933 asset instances in total). This amount and variability of data allow the evaluation of state-of-the-art object detection methods. Also, for five of those assets, there are six different types of defects annotated on an image level (402 defect samples in total, cropped from the UAV images), allowing the evaluation of image classification and unsupervised anomaly detection methods. Thus, we develop a benchmark to serve as a baseline for the community. Their performance, as well as throughput and model size, are assessed. The evaluations show room for improvement in all tasks proposed by InsPLAD. Also, components of power lines from different companies share high similarities, as verified by the datasets cited in the following section. Considering that methods tested in InsPLAD should also perform similarly in power line assets from multiple companies.

To the best of our knowledge, the InsPLAD is the first public dataset for power line asset inspection that offers tasks for each step in a usual inspection pipeline of power line assets: object detection for detecting power line components in UAV images, defect classification in the cropped power line component images, and unsupervised anomaly detection also in cropped asset images. In addition, InsPLAD can be expanded to other \ac{cv} sub-tasks, such as fine-grained classification and small object detection, as it has many images with multiple assets per image. It is intended to spark future research in developing methods for inspection in multiple industries where some defects/faults are not acceptable due to costs or safety concerns, leading to scarcity of data with faulty/anomalous objects, e.g., energy distribution, dangerous spaces, and modern logistics supply chain. In summary, the research gaps from automatic power line visual inspection and the main contributions of InsPLAD to solve them are listed side-by-side in Table \ref{tab:contributions}.

\begin{table}
\caption{Research gap/Contribution pairs addressed by InsPLAD}
\centering
\label{tab:contributions}
\begin{tabular}{p{6.7cm}p{6.7cm}}
\toprule
Research gap    & Contribution     \\ \midrule
Lack of public datasets for power line inspection \cite{nguyen2018automatic} \cite{liu2020data} \cite{xia2022gan}. & A novel dataset for power line assets inspection in the wild, using real-world inspection UAV images. \\
\\ 
Public power line datasets usually do not address multiple vision tasks \cite{tao2020cplid} \cite{tomaszewski2018collection} \cite{lee2017weakly,bian2019monocular}.  & Our dataset can be applied in three vision tasks with deep learning methods: object detection for asset detection and image classification/anomaly detection for defect classification. It also offers multiple computer vision challenges with far more images/defects than other datasets.  \\
\\ 
Power line components present multiple challenging appearances \cite{vieiraesilva2021stnplad}. & InsPLAD presents several categories of power line assets with multiple subtypes, orientations, and scales in uncontrolled environments (in the wild). \\
\\ 
Lack of faulty industrial component imagery taken from real-world scenarios due to costs and safety issues \cite{bergmann2019mvtec}.   & The first public dataset that covers multiple faulty power line assets, allowing the development of supervised and unsupervised methods. \\ 
\\ 
Lack of a power line inspection benchmark to spark new research and select the best setup for real-world scenarios \cite{vieiraesilva2021stnplad} \cite{abdelfattah2020ttpla}. & A comprehensive benchmark with state-of-the-art and popular methods on the proposed dataset to serve as a baseline for future research. \\ \bottomrule
\end{tabular}
\end{table}

In the following section, we present the main public power line datasets available divided by component types and the main methods applied to those. In Section 3 we describe the proposed dataset, showing the data collection protocol, annotation process and its properties. Next, the benchmark is introduced, divided into three parts: Assets Detection, Supervised Fault Classification, and Unsupervised Anomaly Detection. In each one, we show the experiment setup, results, and discussion. Then, InsPLAD's strengths and limitations are presented, followed by the conclusions and future works.

\section{Related Works}


Despite the multiple general-purpose datasets available for object detection \cite{lin2014coco, deng2009imagenet, everingham2010pascal, kuznetsova2020googleopenimages}, image classification \cite{deng2009imagenet, everingham2010pascal}, and anomaly detection \cite{bergmann2019mvtec}, power line inspection condense numerous significant problems for the computer vision community.

The existing public power line datasets only cover some of the various power line components and their defects. They do not have enough data or variability to feed the increasingly data-hungry state-of-the-art methods. Most focus on one or two categories of power line components, mainly Insulators, Transmission Towers, or the conductor itself. In this scenario, InsPLAD presents challenges as multi-scale components, multiple objects per image, and classes with multiple instances of the same component, e.g., Stockbridge Dampers, or underrepresented classes with few samples, e.g., Spacers. It also presents contextual challenges such as cluttered background, uncontrolled outdoor environment, perspective distortion, occlusion, and multiple lighting conditions. Next, the current public power line datasets and the main visual inspection methods for their problems are briefly described.

\subsection{Datasets}

Here, we select the current power line datasets with the following conditions: publicly available data, real-world data, and only data from high-voltage overhead lines. An extensive list of power line datasets is discussed in \cite{ruszczak2023overview}. Table \ref{table:relatedworks} shows a summarised comparison between the {selected datasets and our proposed InsPLAD. 

\subsubsection{Insulator Datasets} 
Tomaszewski et al. \cite{tomaszewski2018collection} propose an \ac{od} Insulator dataset containing 2630 images of size 5616$\times$3744, with one Insulator per image. Despite many samples, the images are repetitive since they come from subsequent frames of nine videos made with a still camera pointing to an Insulator piece suspended by a metal structure. In addition, lighting conditions and the Insulator orientation vary between sets of images. The other Insulator dataset is \ac{cplid} \cite{tao2020cplid}, which is also an OD dataset that consists of 600 real-world images from power line towers and 248 images of segmented, defective Insulators pasted over different backgrounds. The defective Insulators are missing one or more Insulator caps, and the images are of size 1152$\times$864. In it, there are 1,569 annotations, ranging from Insulators and missing caps. 

\subsubsection{Conductor Dataset} The Powerline dataset \cite{lee2017weakly} is a \ac{ss} dataset consisting of 4,200 regular RGB images and 4,200 Infrared images with and without power line conductors. Those were annotated segmenting power lines from the background at a pixel level. Image sizes are 128$\times$128 and 512$\times$512. Within the same context, the VEPL Dataset \cite{canosolis2023vepl} is an SS dataset of vegetation encroachment in power line corridors, a relevant and unexplored topic of power line inspection. Its annotations are the vegetation mask locations in each image.

\subsubsection{Transmission Tower Dataset} The Tower dataset \cite{bian2019monocular} is an OD dataset consisting of 1,300 images of power line transmission towers collected from videos and pictures from the internet in different scenarios. Image size varies for each sample. 

\subsubsection{Multi-category Datasets} \ac{ttpla} \cite{abdelfattah2020ttpla} is an \ac{is} dataset with 1,100 aerial images of size 3840$\times$2160 and two types of power line components: transmission towers and power line conductors. Each power line conductor and transmission tower in every image is annotated at a pixel level as individual instances, totalling 8,987. As for \ac{stnplad} \cite{vieiraesilva2021stnplad}, it consists of 133 aerial images annotated for OD with five power line components: Transmission Tower, Insulator, Tower ID Plate, Yoke, and Stockbridge Damper. The image sizes vary between 5472$\times$3078 or 5472$\times$3648, which allows for a high density of instances per image, making a total of 2,409. It is worth noting that ImageNet \cite{deng2009imagenet} has 1,290 images of transmission towers and power lines with annotated bounding boxes. Most images are not aerial and capture the front of the transmission towers. Finally, PLT-AI Furnas Dataset \cite{oliveira2022pltai} contains 6,295 images with 17,808 annotations for OD. It has five different objects: Insulator, Baliser, Bird Nest, Separator and Stockbridge. Besides Bird Nest, they also have fault data sample annotations, which is quite valuable.}

InsPLAD has 10,607 images, all in 1920$\times$1080 resolution, with more than 28 thousand annotated power line components, categorised into 17 unique classes for the task of Object Detection. Besides that, power line components from five categories were cropped following a squared proportion (equal height and width) and annotated at an image level regarding their operational conditions: whether they are normal or present some defect. This last annotated data was divided into two ways for training and testing: one for a traditional supervised Image Classification (IC) task and the other for an unsupervised Anomaly Detection (AD) task. Finally, InsPLAD contains more images, annotations, components, defects, and vision tasks than all current public power line-related datasets. 

\setlength{\tabcolsep}{1pt}
\begin{table}[h]
\begin{center}
\begin{minipage}{\textwidth}
\caption{Public power line datasets. The defective assets column corresponds to the number of asset categories with defective samples available. Vision tasks: OD - Object Detection; SS - Semantic Segmentation; IS - Instance Segmentation; IC - Image Classification; AD - Anomaly Detection}
\label{table:relatedworks}

\begin{tabular*}{\textwidth}{@{\extracolsep{\fill}}lccccc@{\extracolsep{\fill}}}
\toprule
Dataset                                                               & \makecell{Power line\\asset classes} & Annotations                      & Images                         & \makecell[c]{Defective \\asset classes} & \makecell[c]{Vision tasks}                                                                        
\\
\midrule
\makecell[l]{Power line dataset (2017) 
}                                                     & 1         & 4,200                           & 4,200                           & 0                   & SS                                                                                   
\\
\makecell[l]{Tomaszewski et al. (2018) 
}                                                    & 1         & 2,630 & 2,630 & 0                   & OD                                                                                        
\\
\makecell[l]{Tower dataset (2019) 
} & 1         & Undisclosed                           & 1,300                           & 0                   & OD                                                                                        
\\
CPLID (2020) 
& 1         & 1,569                           & 848                            & 1                   & OD                                                                                        
\\
TTPLA (2020) 
& 2         & 8,987                           & 1,100                           & 0                   & IS                                                                                   
\\
STN PLAD (2021) 
& 5         & 2,409                           & 133                            & 0                   & OD                                                                                        
\\
PLT-AI (2022) 
& 5        & 17,808                           & 6,295                            & 4                   & OD                                                                                        
\\
{\makecell[l]{\textbf{InsPLAD (Ours)}
}}                                                              & \textbf{17}        & \textbf{28,933}                          & \textbf{10,607}                          & \textbf{5}                   & {\textbf{\makecell[c]{OD, IC \& AD}}} 
\\
\bottomrule
\end{tabular*}
\end{minipage}
\end{center}
\end{table}
\setlength{\tabcolsep}{1pt}

\subsection{Inspection methods} \label{insulator_datasets_inspection}

Here, we discuss suitable methods to address the challenges imposed by the datasets described above.

An Object Detection approach is a natural choice to address datasets such as CPLID, Tomaszewski et al., the Tower datasets, and STN PLAD, since their annotations are bounding boxes. 
A straightforward way is to use general-purpose object detection methods, such as \ac{ssd} \cite{liu2016ssd}, \ac{yolo} \cite{redmon2018yolov3}, Faster \ac{rcnn} \cite{ren2015fasterrcnn}, and others. Such object detection datasets would fine-tune those object detectors, which were pre-trained with a large-scale multi-purpose dataset such as MS COCO \cite{lin2014mscoco}. On the other hand, some works propose customised methods to address those datasets. For instance, in the CPLID work \cite{tao2020cplid}, the authors also proposed a method to inspect the annotated insulators with a two-step object detector. It is based on Convolutional Neural Networks (CNNs) to extract features from the images and propose regions of interest, similar to the aforementioned general-purpose object detectors. In the Tower dataset paper, the authors also propose an adaptation of the Faster R-CNN method named Tower R-CNN with fewer convolutional layers to meet their prediction speed and accuracy requirements.

For Semantic Segmentation problems such as in the Powerline dataset, the annotations highlight the pixels in the images representing a particular class, i.e., masks. For instance, the masks in the Powerline dataset represent the precise locations of power line cables. A popular way to address this kind of dataset is to fine-tune general-purpose segmentation methods such as Mask R-CNN \cite{he2017maskrcnn} and U-Net \cite{ronneberger2015unet}. Other works also propose focused techniques for this kind of dataset. For example, in the Powerline dataset \cite{lee2017weakly}, the authors also propose a method of inspecting the power lines by classifying sub-regions using a sliding window and a CNN. Specifically, a sub-region is filtered out if it is classified into an image without any power line. In that case, if a sub-region is classified into an image containing a power line, then its feature maps of intermediate convolutional layers are combined to visualise the location of the power line.

For Instance Segmentation, there are also general-purpose methods. For example, the authors of TTPLA use the Yolact \cite{bolya2019yolact} method for real-time instance segmentation with different backbones to create a baseline for that dataset. Other general-purpose instance segmentation methods, such as Mask R-CNN and CenterMask \cite{lee2020centermask}, could also be applied.

Our proposed dataset, InsPLAD, has three different types of annotations. Bounding boxes for Object Detection and Image-level labels for Image Classification and Anomaly Detection. Object Detection methods were already presented. 
Regarding Image Classification, CNNs, e.g., ResNet \cite{he2016deep} and EfficientNet \cite{tan2019efficientnet}, are still quite popular to approach this problem. Other recently popular methods for image classification are the ones based on Vision Transformers, such as the Swin-Transformer \cite{liu2021swin}. \ac{fcn}, as \ac{mlp}, still present results compatible with state-of-the-art, as seen in \cite{tolstikhin2021mlp}.

Finally, Unsupervised Anomaly Detection is a recent research topic in Computer Vision. Methods to address it are usually based on Autoencoders \cite{bergman2019improving}, \ac{gan} \cite{zaheer2020old}, and Normalizing flows \cite{rudolph2021same, rudolph2022fully}, the last one being more recent and showing promising results.  

\section{Dataset Description}
\label{sec:dataset}
This section describes the InsPLAD dataset and its divisions for object detection, fault classification, and anomaly detection. Initially, the process of capturing images in active power transmission lines using drones is described in Subsection \ref{sec:protocol}. Next, the Subsection \ref{sec:annotation} discusses the annotation process of the images. Finally, in Subsection \ref{sec:properties}, the generated datasets are presented and discussed.

\subsection{Data Collection Protocol}
\label{sec:protocol}

Before starting the data collection, the main assets that may present problems were defined in cooperation with a power line company. The captured overline power line is a high-voltage 500 kV transmission line from the northeast region of Brazil. From the collaboration, 17 power line components were selected: Spiral Damper, Stockbridge Damper, Glass Insulator, Glass Insulator's Big Shackle, Glass Insulator's Small Shackle, Glass Insulator's Tower Shackle, Lightning Rod's Shackle, Lightning Rod's Suspension, Tower ID Plate, Polymer Insulator, Polymer Insulator's Lower Shackle, Polymer Insulator's Upper Shackle, Polymer Insulator's Tower Shackle, Spacer, Vari-grip, Yoke, Yoke's Suspension.

After defining the assets that would receive attention during data collection, a capture protocol was developed to ensure the standardization of captured photos. The protocol aims to ensure that all highlighted components have images centring and standardise how the images should be captured to improve the inspection process. After that definition, the data collection protocol was implemented and refined between October and December 2020, using the DJI Matrice 210 V2 drone coupled to the Zenmuse z30 camera, which produces images with Full HD resolution (1920$\times$1080). Two qualified drone pilots participated in a training session with researchers to follow the capture protocol. The pilots then captured the images under the supervision of the researchers to make last-minute adjustments to the protocol. Images of 226 power transmission line towers were captured during the process. Towers vary between structural types and each asset class contained in the tower. A total of 10,607 images were captured that focused on the selected assets. 

The captured power line has three phases, A, B, and C, with B being the middle one. Phases A and C are typically mirrored as they are on opposite sides. This protocol was studied, improved, and refined daily during the collection phase until the latest version was reached. The detailed final protocol is presented below.

\begin{enumerate}


\item Photo of Tower ID Plate: One angle - One image (10 to 15 m away)

\begin{figure}[H]
\centering
\label{fig:plate}
\includegraphics[width=0.4\textwidth]{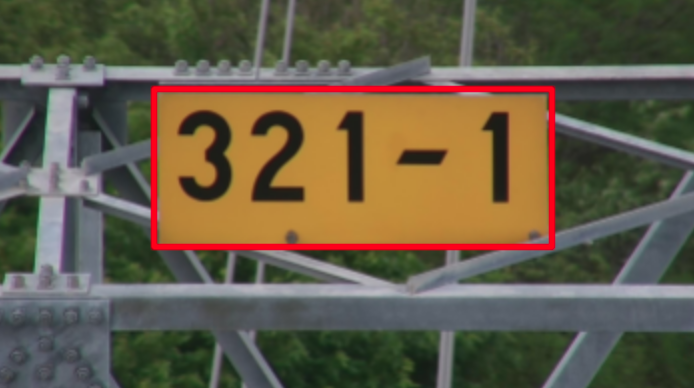}
\caption{ID Plate}
\end{figure}

\item PHASE A

\begin{enumerate}
\item Stockbridge Damper set: One angle, framing on the tower's side - One image (20 to 25 m)

\begin{figure}[H]
\centering
\label{fig:stock}
\includegraphics[width=0.4\textwidth]{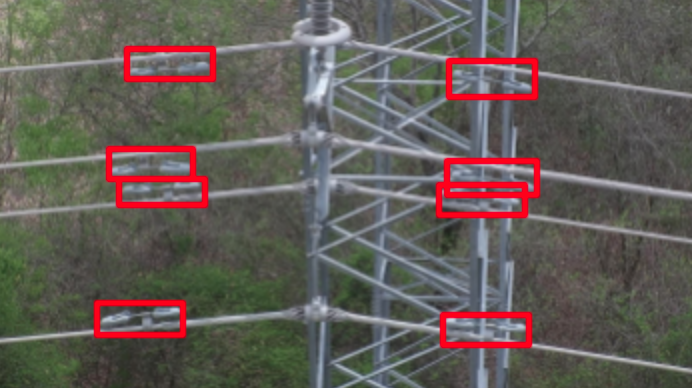}
\caption{Stockbridge Damper}
\end{figure}

\item Yoke Suspension: One angle, framing on the tower's side - Two images, One per suspension pair (10 m)

\begin{figure}[H]
\centering
\label{fig:yokesus}
\includegraphics[width=0.4\textwidth]{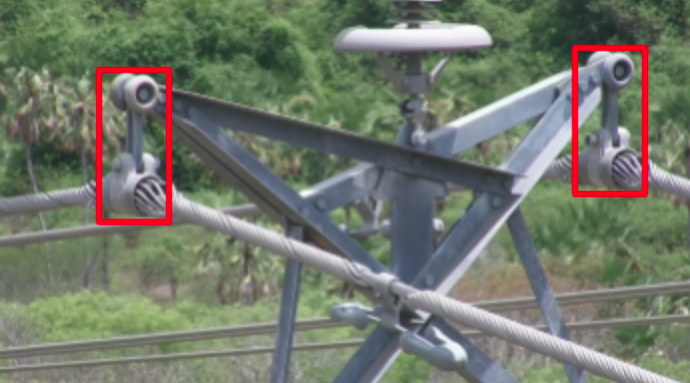}
\caption{Yoke Suspension}
\end{figure}

\item Yoke: One angle, framing on the tower's side - One image (15 m)

\begin{figure}[H]
\centering
\label{fig:yoke}
\includegraphics[width=0.4\textwidth]{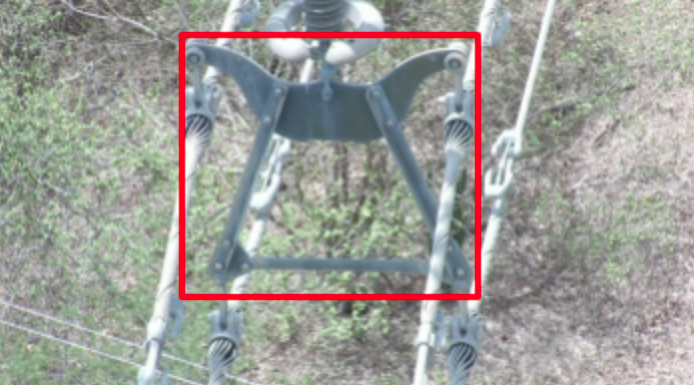}
\caption{Yoke}
\end{figure}

\item Polymer/Glass Insulator Lower/Small Shackle: One angle, framing on the tower's side - One image (20 m)

\begin{figure}[H]
\centering
\label{fig:polymerls}
\includegraphics[width=0.4\textwidth]{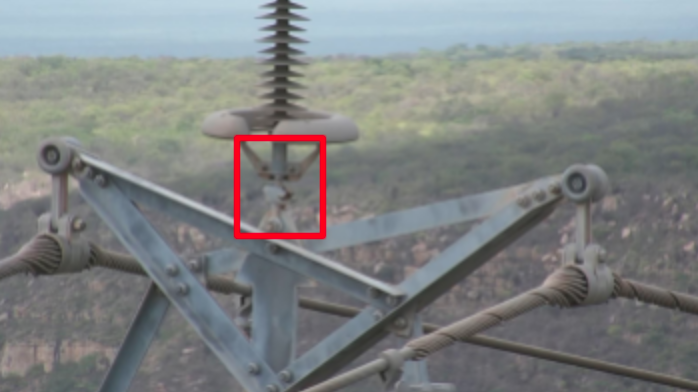}
\caption{Polymer Insulator Lower Shackle}
\end{figure}

\item Polymer/Glass Insulator:

\begin{enumerate}

\item An angle, wide frame on the tower's side to show which Insulator will be detailed - One image (15 m)

\item One angle, strict framing in the body of the Insulator - Four images, i.e. the Insulator is divided into at least Four images (25 m)

\begin{figure}[H]
\centering
\label{fig:polymer}
\includegraphics[width=0.4\textwidth]{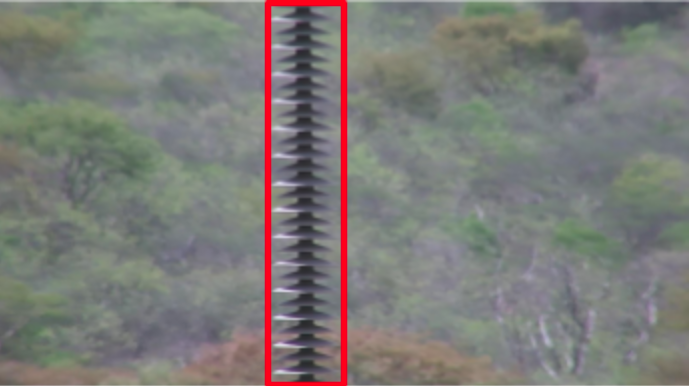}
\caption{Polymer Insulator}
\end{figure}

\end{enumerate}

\item Polymer/Glass Insulator Upper/Big Shackle: One angle, side framing - One image (10 m)

\begin{figure}[H]
\centering
\label{fig:polymerus}
\includegraphics[width=0.4\textwidth]{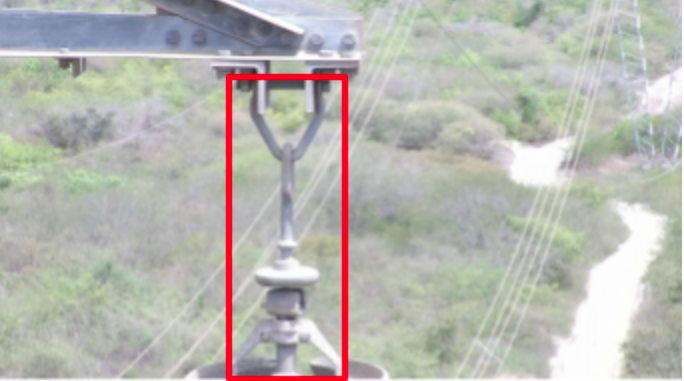}
\caption{Polymer Insulator Upper Shackle}
\end{figure}

\end{enumerate}

\item Vari-grip on the left (if the tower has Vari-grips): One angle, diagonal framing - One image (9 m)

\begin{figure}[H]
\centering
\label{fig:varigrip}
\includegraphics[width=0.4\textwidth]{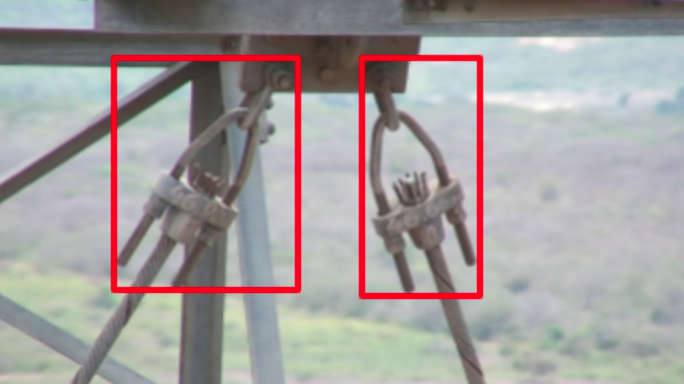}
\caption{Vari-grip}
\end{figure}

\item Lightning Rod Suspension/Shackle

\begin{figure}[H]
\centering
\label{fig:rodsus}
\includegraphics[width=0.4\textwidth]{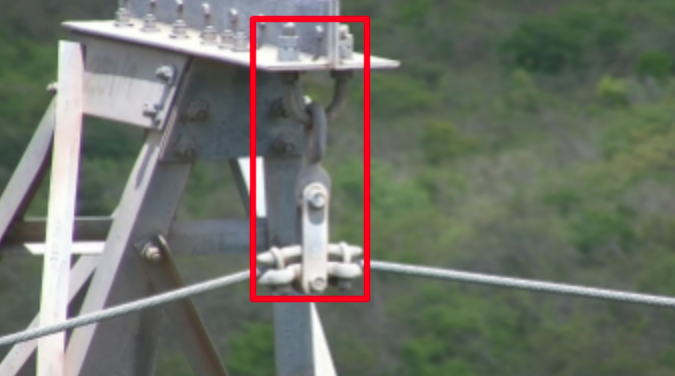}
\caption{Lightning Rod Suspension}
\end{figure}

\begin{enumerate}
\item Farther Lightning Rod Suspension/Shackle (opposite side): One angle, diagonal framing - One image (15 m)

\item Closest Lightning Rod Suspension/Shackle: One angle, side framing - One image (15 m)
\end{enumerate}

\item PHASE B: Repeat the same procedure as in PHASE A, only with the drone positioned in the front frame between the tower Lightning Rods. Distances are the same as in PHASE A: only change the sides but keep the angles on the opposite side.

\item Vari-grip on the right side: One angle, diagonal framing - One image (10 m)

\item PHASE C: Repeat the same procedure as in PHASE A, only now with the drone on the other side of the tower. Again, distances are the same as in PHASE A: The Vari-grip can be captured in PHASE A and PHASE C, always from the side to maintain safety, seeking a better angle with zoom and brightness adjustments.
Repeat Step 5, only now positioning the drone on the other side of the tower. Again, distances are the same as in PHASE A.

\end{enumerate}

\subsection{Annotation Process}
\label{sec:annotation}

\begin{figure*}
\centering
\begin{subfigure}[b]{0.3\textwidth}
\includegraphics[width=\linewidth]{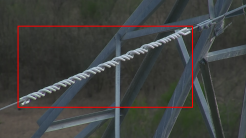}
\caption{}
\end{subfigure}
\hfill
\begin{subfigure}[b]{0.3\textwidth}
\includegraphics[width=\linewidth]{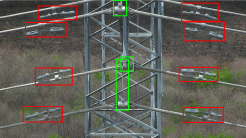} 
\caption{} 
\end{subfigure}
\hfill
\begin{subfigure}[b]{0.3\textwidth}
\includegraphics[width=\linewidth]{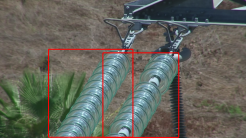} 
\caption{} 
\end{subfigure}

\begin{subfigure}[b]{0.3\textwidth}
\includegraphics[width=\linewidth]{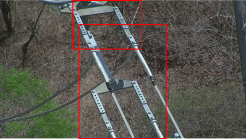} 
\caption{} 
\end{subfigure}
\hfill
\begin{subfigure}[b]{0.3\textwidth}
\includegraphics[width=\linewidth]{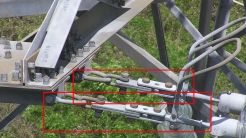} 
\caption{} 
\end{subfigure}
\hfill
\begin{subfigure}[b]{0.3\textwidth}
\includegraphics[width=\linewidth]{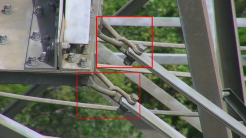}
\caption{} 
\end{subfigure}

\centering
\begin{subfigure}[b]{0.3\textwidth}
\includegraphics[width=\linewidth]{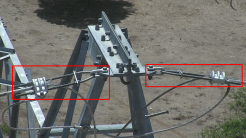}
\caption{} 
\end{subfigure}
\hfill
\begin{subfigure}[b]{0.3\textwidth}
\includegraphics[width=\linewidth]{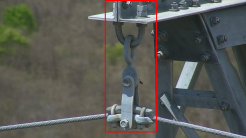} 
\caption{} 
\end{subfigure}
\hfill
\begin{subfigure}[b]{0.3\textwidth}
\includegraphics[width=\linewidth]{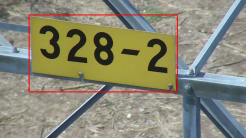} 
\caption{} 
\end{subfigure}

\begin{subfigure}[b]{0.3\textwidth}
\includegraphics[width=\linewidth]{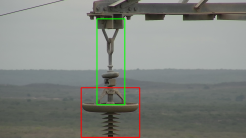} 
\caption{} 
\end{subfigure}
\hfill
\begin{subfigure}[b]{0.3\textwidth}
\includegraphics[width=\linewidth]{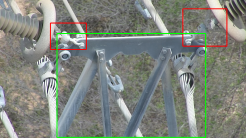} 
\caption{} 
\end{subfigure}
\hfill
\begin{subfigure}[b]{0.3\textwidth}
\includegraphics[width=\linewidth]{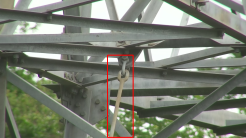} 
\caption{} 
\end{subfigure}

\begin{subfigure}[b]{0.3\textwidth}
\includegraphics[width=\linewidth]{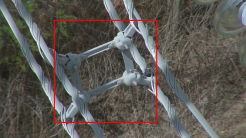} 
\caption{} 
\end{subfigure}
\hspace{25pt}
\begin{subfigure}[b]{0.3\textwidth}
\includegraphics[width=\linewidth]{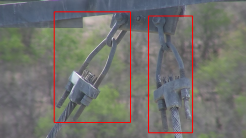} 
\caption{} 
\end{subfigure}

\caption{All 17 assets of the InsPLAD-det. [R] and [G] indicate the colours of the bounding box of that asset in the image, [R] being Red and [G] Green. The assets are: (a) Damper - Spiral [R], (b) Damper - Stockbridge [R] and Yoke Suspension [G], (c) Glass Insulator [R], (d) Glass Insulator Big Shackle [R], (e) Glass Insulator Small Shackle [R], (f) Glass Insulator Tower Shackle [R], (g) Lightning Rod Shackle [R], (h) Lightning Rod Suspension [R], (i) Tower ID Plate [R], (j) Polymer Insulator [R] and Polymer Insulator Upper Shackle [G], (k) Polymer Insulator Lower Shackle [R] and Yoke [G], (l) Polymer Insulator Tower Shackle [R], (m) Spacer [R], (n) Vari-grip [R]. Best seen in colour}
\label{fig:insplad-det}
\end{figure*}

The data was annotated by two annotators who received instructions from specialists about the names of the assets and regions of interest, which should be contained in the detection bounding boxes. Those bounding boxes are already presented in the data collection protocol above. In addition, all annotators were in constant contact with each other to ensure standardization in the data annotation process. The bounding box annotations were made with the help of the LabelImg tool \cite{tzutalin2015labelimg}. This part of the dataset was named InsPLAD-det. All 17 asset classes and samples from InsPLAD-det can be seen in Figure \ref{fig:insplad-det}.

After creating InsPLAD-det, it was noticed that some asset classes had many faults, even for an active power line. An annotation process for the asset conditions with relevant defective data was carried out for this. These annotations enable the application of methods that seek to identify defects in the assets of power lines. With annotations regarding the state of the assets, it is possible to use image classification techniques to classify them according to their condition and point out defects. The same annotators from InsPLAD-det conducted the process. They also met constantly to ensure standardization in the annotation process. 

\begin{figure*}
\centering
\begin{subfigure}[b]{0.19\textwidth}
\includegraphics[width=\linewidth]{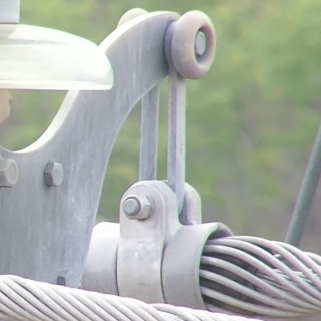}
\end{subfigure}
\begin{subfigure}[b]{0.19\textwidth}
\includegraphics[width=\linewidth]{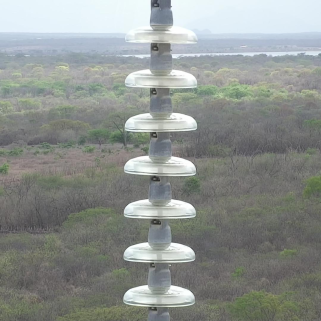} 
\end{subfigure}
\begin{subfigure}[b]{0.19\textwidth}
\includegraphics[width=\linewidth]{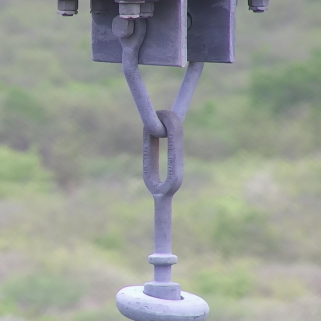} 
\end{subfigure}
\begin{subfigure}[b]{0.19\textwidth}
\includegraphics[width=\linewidth]{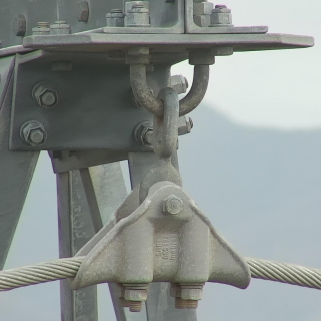} 
\end{subfigure}
\begin{subfigure}[b]{0.19\textwidth}
\includegraphics[width=\linewidth]{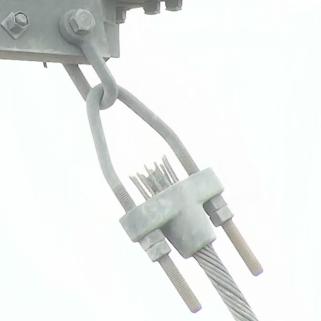} 
\end{subfigure}

\begin{subfigure}[b]{0.19\textwidth}
\includegraphics[width=\linewidth]{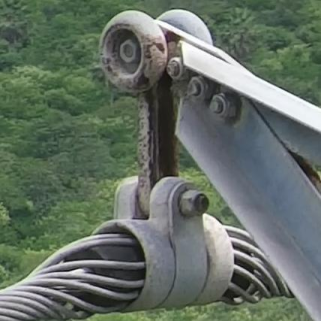}
\end{subfigure}
\begin{subfigure}[b]{0.19\textwidth}
\includegraphics[width=\linewidth]{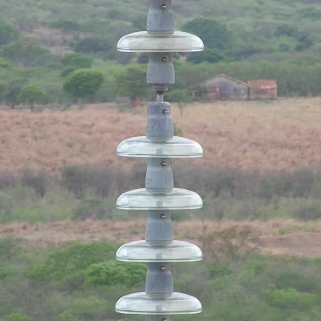} 
\end{subfigure}
\begin{subfigure}[b]{0.19\textwidth}
\includegraphics[width=\linewidth]{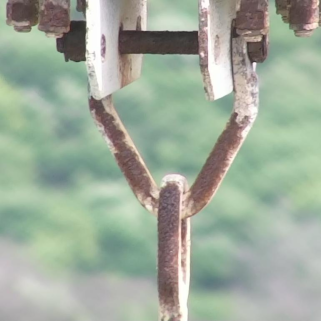} 
\end{subfigure}
\begin{subfigure}[b]{0.19\textwidth}
\includegraphics[width=\linewidth]{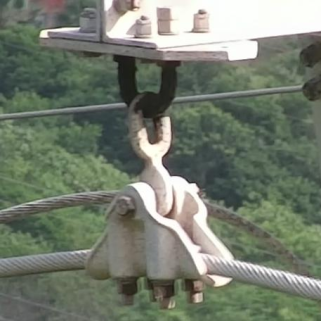} 
\end{subfigure}
\begin{subfigure}[b]{0.19\textwidth}
\includegraphics[width=\linewidth]{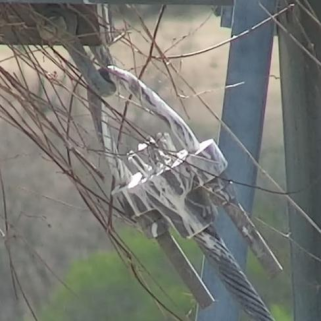} 
\end{subfigure}

\caption{All five asset classes of InsPLAD-fault. The first row contains images of the assets in their normal state, and the second contains the defective assets. From left to right, the assets in the images are Yoke Suspension (normal/rusty), Glass Insulator (normal/missing cap), Polymer Insulator Upper Shackle (normal/rusty), Lightning Rod Suspension (normal/rusty), and Vari-grip (normal/bird's nest). Vari-grip can also be rusty}
\label{fig:insplad-fault}
\end{figure*}

The detected assets were cut from the original image using the regions of interest information contained in the InsPLAD-det's bounding boxes to annotate the assets' states. Cropped from the original images, the new images were grouped by asset class before starting the classification step regarding their status. The Pigeon tool \cite{germanidis2017pigeon} was used to help annotate the assets instances states. For each image analysed, the annotators classified its state as normal or the name of the defect found. With this process completed, the InsPLAD-fault dataset was created. InsPLAD-fault assets are shown in Figure \ref{fig:insplad-fault}. It shows the five asset classes that make up the dataset. In the first row, there are samples of assets under normal conditions, and in the second, defective samples. 

\subsection{Properties}
\label{sec:properties}

After building the InsPLAD-det detection dataset, it was possible to obtain the dataset analytics. Table \ref{tab:insplad-det} shows essential dataset details that include the names of all assets, the total number of images containing each asset, the total number of annotation instances for each asset, and the average size of the annotations. It is possible to notice a total of 28,933 annotations in the dataset, a large number compared with other related datasets \cite{lee2017weakly,tomaszewski2018collection,tao2020cplid,bian2019monocular,abdelfattah2020ttpla,vieiraesilva2021stnplad}.

The images from each transmission tower were grouped during the data split to avoid similar images, e.g., the same object from different angles, in the InsPLAD-det training and test sets. Thus, 80\% of the transmission towers were selected for training and 20\% for testing. Also, the division was done carefully to evenly distribute each tower type and asset class between training and testing.

\setlength{\tabcolsep}{1pt}
\begin{table}[ht]
\begin{center}
\begin{minipage}{\textwidth}
\caption{Description of InsPLAD-det. The table shows the 17 assets in InsPLAD-det, the number of images containing each asset, the total number of annotations for each asset, the average size of annotations in squared pixels, and the division of training/test samples}
\label{tab:insplad-det}
\begin{tabular*}{\textwidth}{@{\extracolsep{\fill}}lcccc@{\extracolsep{\fill}}}
\toprule
\multicolumn{1}{l}{\begin{tabular}[l]{@{}l@{}}Asset category\end{tabular}} & \begin{tabular}[c]{@{}c@{}}Number of\\ images\end{tabular} & \begin{tabular}[c]{@{}c@{}}Number of\\ annotations\end{tabular} & \begin{tabular}[c]{@{}c@{}}Average annotation\\ size (px²)\end{tabular} & \begin{tabular}[c]{@{}c@{}}Train/test\\ samples\end{tabular} 
\\
\midrule
\begin{tabular}[c]{@{}l@{}}Damper - Spiral\end{tabular}                             & 943                                                                 & 1020                                                                     & 518.12 $\pm$ 192.73        & 831/189                                                                 \\              
\begin{tabular}[c]{@{}l@{}}Damper - Stockbridge\end{tabular}                        & 1761                                                                & 6953                                                                     & 162.12                                                                             $\pm$ 58.25                                    & 5699/1254                                      \\
\begin{tabular}[c]{@{}l@{}}Glass Insulator\end{tabular}                             & 2778                                                                & 2978                                                                     & 546.08                                                                             $\pm$ 206.05                                 & 2015/963                                        \\
\begin{tabular}[c]{@{}l@{}}Glass Insulator Big Shackle\end{tabular}                   & 152                                                                 & 259                                                                      & 473.10                                                                             $\pm$ 186.35                               & 110/149                                          \\
\begin{tabular}[c]{@{}l@{}}Glass Insulator Small Shackle\end{tabular}                 & 143                                                                 & 263                                                                      & 361.12                                                                             $\pm$ 181.44                                & 128/135                                          \\
\begin{tabular}[c]{@{}l@{}}Glass Insulator Tower Shackle\end{tabular}                 & 106                                                                 & 195                                                                      & 204.29                                                                             $\pm$ 82.68                                 & 98/97                                         \\
\begin{tabular}[c]{@{}l@{}}Lightning Rod Shackle\end{tabular}                       & 112                                                                 & 195                                                                      & 286.54                                                                             $\pm$ 110.28                                     & 170/25                                    \\
\begin{tabular}[c]{@{}l@{}}Lightning Rod Suspension\end{tabular}                    & 709                                                                 & 710                                                                      & 363.26                                                                             $\pm$ 166.37                                  & 618/92                                       \\
\begin{tabular}[c]{@{}l@{}}Tower ID Plate\end{tabular}                              & 242                                                                 & 242                                                                      & 611.30                                                                             $\pm$ 166.86                                   & 198/44                                      \\
\begin{tabular}[c]{@{}l@{}}Polymer Insulator\end{tabular}                           & 3173                                                                & 3244                                                                     & 534.65                                                                             $\pm$ 233.90                                  & 2389/855                                       \\
\begin{tabular}[c]{@{}l@{}}Pol. Insulator Lower Shackle\end{tabular}                & 1760                                                                & 1842                                                                     & 153.21                                                                             $\pm$ 86.44                                 & 1460/382                                         \\
\begin{tabular}[c]{@{}l@{}}Pol. Insulator Upper Shackle\end{tabular}                & 1691                                                                & 1692                                                                     & 392.55                                                                             $\pm$ 160.92                                & 1315/377                                         \\
\begin{tabular}[c]{@{}l@{}}Pol. Insulator Tower Shackle\end{tabular}                & 57                                                                  & 57                                                                       & 182.13                                                                             $\pm$ 58.87                                & 47/10                                          \\
Spacer                                                                                & 93                                                                  & 94                                                                       & 463.99                                                                             $\pm$ 182.75                                   & 72/22                                      \\
Vari-grip                                                                             & 560                                                                 & 1008                                                                     & 461.75                                                                             $\pm$ 139.54                              & 846/162                                           \\
Yoke                                                                                  & 1661                                                                & 1661                                                                     & 753.20                                                                             $\pm$ 290.48                              & 1343/318                                           \\
Yoke Suspension                                                                            & 2716                                                                & 6520                                                                     & 221.11                                                                             $\pm$ 110.90    & 5270/1250 \\ \bottomrule                                                                   
\end{tabular*}
\end{minipage}
\end{center}
\end{table}
\setlength{\tabcolsep}{1pt}

From the InsPLAD-det, it was possible to build the InsPLAD-fault with the classes that defects were observed. InsPLAD-fault has the properties shown in Table \ref{tab:insplad-fault}. Unlike InsPLAD-det, InsPLAD-fault has five component classes, since not all InsPLAD-det's 17 classes had sufficient samples -- the majority had no sample at all --  of defective components. InsPLAD-fault's five classes are divided into a set of normal and defective images. For Lightning Rod Suspension, Polymer Insulator Upper Shackle, Vari-grip, and Yoke classes, samples with rusty components were found. For the Glass Insulator class, there were missing cap faults. Additionally, for the Vari-grip, the presence of a bird's nest was also found in some samples, so it is the only class with two types of faults noted. Figure \ref{fig:insplad-fault} shows the examples of defects mentioned.

Due to the low number of faults in power lines, an unsupervised anomaly detection approach is also adopted to detect faults. Therefore, InsPLAD-fault was adapted into two approaches: supervised fault classification and unsupervised anomaly detection. The data was split to have sample representativeness in both training and testing sets and minimise the impact of unbalance between the normal and faulty classes. For this, only a smaller set of normal images was selected to compose the training set of assets under normal conditions. Table \ref{tab:insplad-fault} shows the data split for both approaches. The main difference between both approaches is that, for anomaly detection, defective examples do not appear in the training set, only in the test set. Therefore, having more normal examples in anomaly detection training does not imply class imbalance. The other difference is that, for the fault classification approach, the faulty samples of the training sets were augmented by a factor of ten using traditional data augmentation techniques through the Albumentations library \cite{buslaev2020albumentations}. This is done to produce balanced training sets.

\setlength{\tabcolsep}{1pt}
\begin{table}
\begin{center}
\begin{minipage}{\textwidth}
\caption{Total images and distribution for training and testing for both InsPLAD-fault approaches. All assets contain only one fault/anomaly type except the Vari-grip, which had its faults divided into two categories separated by a forward slash (bird's nest/rust)}
\label{tab:insplad-fault}
\begin{tabular*}{\textwidth}{@{\extracolsep{\fill}}lccccccc@{\extracolsep{\fill}}}
\toprule
\multirow{4}{*}{Asset category} & \multicolumn{4}{c}{Fault classification}               & \multicolumn{3}{c}{Anomaly detection}     \\ \cmidrule(lr){2-5} \cmidrule(lr){6-8}
                                & \multicolumn{2}{c}{Train} & \multicolumn{2}{c}{Test} & \multicolumn{1}{c}{Train} & \multicolumn{2}{c}{Test}      \\ \cmidrule(lr){2-3} \cmidrule(lr){4-5} \cmidrule(lr){6-6} \cmidrule(lr){7-8}
                                & Normal        & Fault        & Normal        & Fault       & Normal        & Normal        & Anomaly         \\ 
\midrule
Glass Insulator                 & 691   & 60          & 29                 & 30         & 2298         & 581         & 90           \\
Lightning Rod Suspension        & 348   & 30          & 231                & 20         & 462          & 117         & 50           \\
\makecell[l]{Pol. Insulator Upper Shackle}    & 742   & 48          & 31                 & 33         & 935          & 235         & 102          \\
Vari-grip                       & 358   & 48/28       & 238               & 23/20      & 477          & 114         & 63/48        \\ 
Yoke Suspension                 & 299         & 29    & 5742               & 20         & 4834         & 1207        & 49           \\
\bottomrule
\end{tabular*}
\end{minipage}
\end{center}
\end{table}
\setlength{\tabcolsep}{1pt}

\section{Benchmark}

We conducted several experiments with state-of-the-art methods for object detection, image classification, and anomaly detection to serve as a benchmark on InsPLAD. It is intended to provide a baseline for future work. The experiments were conducted in the Ubuntu 20.04 OS. For the ones in Subections~\ref{sec:assetdet} and \ref{sec:fault} - Supervised fault classification, an Nvidia RTX 3080Ti GPU was used, and in \ref{sec:anomaly} - Unsupervised anomaly detection, an Nvidia RTX 2080Ti was used.

\subsection{InsPLAD-det: Asset Detection} 
\label{sec:assetdet}

The first challenge imposed by our dataset is detecting power line assets in UAV-captured RGB images, which is an object detection task. We fine-tune popular and state-of-the-art object detectors, namely Faster R-CNN \cite{ren2015fasterrcnn}, SSD \cite{liu2016ssd}, YOLOv3 \cite{redmon2018yolov3}, RetinaNet \cite{lin2017focal}, Cascade R-CNN \cite{cai2018cascade}, TOOD \cite{feng2021tood} and DetectoRS \cite{qiao2021detectors}.We also choose to experiment with older methods, such as YOLOv3, Faster R-CNN and SSD, because they are popular in power line inspection research \cite{liu2020data}.

\subsubsection{Setup}


All object detection methods were evaluated using their implementations from the MMDetection toolbox by OpenMMLab \cite{mmdetection}, based on the PyTorch framework \cite{paszke2019pytorch}, and publicly available on GitHub \cite{MMDetection_Contributors_OpenMMLab_Detection_Toolbox_2018}. They were initialised from publicly available pre-trained weights provided by the MMDetection toolbox and fine-tuned until loss stabilization. The methods were trained in their standard configurations of parameters and hyperparameters. This is summarised in Table~\ref{tab:hyperod}.
\setlength{\tabcolsep}{3pt}
\begin{table}
\caption{Experimental configurations of the object detection methods}
\centering
\label{tab:hyperod}
\begin{tabular}{@{}llcccc@{}}
\toprule
Method    & Backbone & \makecell[c]{Input size} & Epochs & \makecell[c]{Batch size} & \makecell[c]{Learning Rate}     \\ \midrule
SSD512           & VGG16 & $512\times512$    & 12    & 64    & 0.002 \\
RetinaNet     & \makecell[l]{ResNeXt-101} & $1333\times800$      & 12    & 16    & 0.01  \\
YOLOv3        & \makecell[l]{Darknet-53 608} & $608\times608$    & 12    & 64    & 0.001 \\
TOOD          & \makecell[l]{ResNet-101} & $1333\times800$    & 24    & 16    & 0.01 \\
\makecell[l]{Faster R-CNN}  & \makecell[l]{ResNet-101} & $1333\times800$    & 12    & 16    & 0.02 \\
\makecell[l]{Cascade R-CNN} & \makecell[l]{ResNeXt-101} & $1333\times800$    & 12    & 16    & 0.02 \\
DetectoRS     & \makecell[l]{Cascade + ResNet-50} & $1333\times800$    & 12    & 16    & 0.02 \\ \bottomrule
\end{tabular}
\end{table}
\setlength{\tabcolsep}{3pt}

Finally, the best-evaluated checkpoint was selected for further comparison by the end of each training.

The primary metric used for comparison is the Box AP from MS COCO, also known as AP (with \ac{iou} as $0.50:0.95$), currently the main object detection metric. AP$^{50}$ and AP$^{75}$ are used as secondary metrics. We also present weight size in megabytes and throughput, the latter being the average number of inferences the model makes per second during the test. This last information is relevant considering a real-time inspection scenario or a scenario with limited resources.  

\subsubsection{Results and Discussion}

Firstly, Table~\ref{tab:odperclass} shows the Box AP results for each object detector detailed per asset category. The variations between class AP values are due to object complexity, object size relative to the original image size from which it has been cropped, and the amount of training and testing samples. In addition, these variations could be responsible for the absence of an obvious lead detector for all classes, which can be further investigated.  

\setlength{\tabcolsep}{1pt}
\begin{table}
\begin{center}
\begin{minipage}{\textwidth}
\caption{Box AP for each of the seventeen asset categories in every experimented object detection method. The best results for each class are highlighted in bold}
\label{tab:odperclass}
\resizebox{\textwidth}{!}{\begin{tabular*}{\textwidth}{@{\extracolsep{\fill}}lccccccc@{\extracolsep{\fill}}}
\toprule
\makecell[l]{Asset category}           & \makecell[c]{SSD} & \makecell[c]{RetinaNet} & \makecell[c]{YOLOv3} & \makecell[c]{TOOD} & \makecell[c]{Faster\\R-CNN} & \makecell[c]{Cascade\\R-CNN} & \makecell[c]{DetectoRS} \\
\midrule
\makecell[l]{Damper - Spiral}               & 0.870                      & 0.862                                           & 0.650                               & \textbf{0.959}                         & 0.895                                        & 0.918                                               & 0.945                                        \\
\makecell[l]{Damper - Stockbridge}          & 0.815                      & 0.845                                           & 0.701                               & \textbf{0.857}                         & 0.835                                        & 0.842                                               & 0.848                                        \\
\makecell[l]{Glass Insulator}             & 0.803                      & 0.838                                           & 0.752                               & 0.889                                  & 0.840                                        & 0.880                                               & \textbf{0.893}                               \\
\makecell[l]{Gl. Ins. Big Shackle}   & \textbf{0.320}                      & 0.226                                           & 0.0012                               & 0.184                                  & 0.137                                        & 0.203                                               & 0.248                               \\
\makecell[l]{Gl. Ins. Small Shackle} & 0.0016                      & 0.275                                           & 0.189                               & 0.214                                  & 0.264                                        & \textbf{0.280}                                      & 0.270                                        \\
\makecell[l]{Gl. Ins. Tower Shackle} & 0.377                      & 0.373                                           & 0.246                               & 0.392                                  & 0.360                                        & \textbf{0.433}                                      & 0.413                                        \\
\makecell[l]{Lightning Rod Shackle}          & 0.566                      & 0.547                                           & 0.402                               & 0.569                                  & 0.490                                        & 0.561                                               & \textbf{0.595}                               \\
\makecell[l]{Lightning Rod Susp.}       & 0.869                      & \textbf{0.928}                                  & 0.774                               & \textbf{0.928}                                  & 0.911                                        & 0.914                                               & 0.911                                        \\
\makecell[l]{Tower ID Plate}              & 0.967                      & 0.983                                           & 0.725                               & 0.984                                  & 0.952                                        & 0.978                                               & \textbf{0.990}                               \\
\makecell[l]{Polymer Insulator}           & 0.850                      & 0.953                                           & 0.785                               & 0.951                                  & 0.921                                        & 0.950                                               & \textbf{0.954}                               \\
\makecell[l]{Pol. Ins. Lower Shackle} & 0.578                      & 0.639                                           & 0.453                               & 0.637                                  & 0.611                                        & 0.644                                               & \textbf{0.648}                               \\
\makecell[l]{Pol. Ins. Upper Shackle} & 0.796                      & 0.855                                           & 0.715                               & \textbf{0.872}                         & 0.835                                        & 0.846                                               & 0.857                                        \\
\makecell[l]{Pol. Ins. Tower Shackle} & 0.507                      & \textbf{0.531}                                           & 0.377                               & 0.371                                  & 0.500                                        & 0.405                                               & 0.528                               \\
Spacer                                      & 0.332                      & 0.486                                           & 0.083                               & 0.367                                  & 0.410                                        & 0.456                                               & \textbf{0.487}                               \\
Vari-grip                                   & 0.914                      & 0.948                                           & 0.826                               & \textbf{0.954}                         & 0.930                                        & 0.941                                               & \textbf{0.954}                                        \\
Yoke                                        & 0.832                      & 0.866                                           & 0.676                               & \textbf{0.880}                         & 0.855                                        & \textbf{0.880}                                               & 0.864                                        \\
Yoke Suspension                                  & 0.808                      & 0.853                                           & 0.769                               & \textbf{0.856}                         & 0.842                                        & 0.855                                               & 0.855                                        \\ \midrule
{Average}                            & 0.674                      & 0.706                                           & 0.546                               & 0.698                                  & 0.682                                        & 0.705                                               & \textbf{0.721}                               \\ \bottomrule
\end{tabular*}}
\end{minipage}
\end{center}
\end{table}
\setlength{\tabcolsep}{1pt}

Table~\ref{table:odpermethod} summarises the performance of all object detectors. DetectoRS achieves the best performance considering Box AP and AP$^{75}$ metrics, with a 0.721 Box AP and a 0.749 AP$^{75}$; RetinaNet reaches second best in both of them, while it achieves the best result related to AP$^{50}$ with 0.891; in this case, DetectoRS reaches second best as well as SSD and Cascade R-CNN. DetectoRS achieving the best overall performance is within expectations since it is a state-of-the-art multi-stage object detector. However, the performance of RetinaNet is the second-best overall, surpassing newer one- and multi-stage techniques.

On the other hand, SSD produces the fastest and most lightweight model, with 48.3 inferences per second during the test. Regarding performance, surprisingly, SSD reaches the same result as DectectoRS in terms of AP$^{50}$, presenting a high potential to be used in real-time power line inspections due to its good performance and fast inference compared to the other methods.

\setlength{\tabcolsep}{2pt}
\begin{table}
\begin{center}
\caption{Performance, weight size in megabytes, and throughput results from state-of-the-art object detection methods. The first four methods correspond to one-stage object detection methods, while the last three are multi-stage approaches. Results in bold are the best for each aspect}
\label{table:odpermethod}
\begin{tabular}{lccccccc}
\toprule
    Method                        & \multicolumn{1}{c}{Reference} & \makecell[c]{Weights size \\(MB)} & \multicolumn{1}{c}{Inferences s$^{-1}$} & \makecell[c]{N° of \\parameters} & \multicolumn{1}{c}{Box AP} & \multicolumn{1}{l}{AP$^{50}$} & \multicolumn{1}{l}{AP$^{75}$} \\
\midrule
SSD 
& ECCV16         & \textbf{215.2}    & \textbf{48.3}  & \textbf{36M} & 0.674              & 0.885             & 0.712             \\
RetinaNet 
& ICCV17         & 756.2             & 11.7           & 56.3M & 0.706              & \textbf{0.891}    & 0.741             \\
YOLOv3 
& arXiv18        & 493.3             & 41.6           & 61.9M & 0.546              & 0.825             & 0.603             \\
TOOD 
& ICCV21         & 427.7             & 13.2           & 53.3M & 0.698              & 0.861             & 0.718             \\ 
Faster R-CNN 
& NeurIPS15      & 483.3             & 18.2           & 60.5M & 0.682              & 0.874             & 0.723             \\
Cascade R-CNN 
& CVPR18         & 704.3             & 14.0           & 87.8M & 0.705              & 0.885             & 0.733             \\
DetectoRS 
& CVPR21         & 990.9             & 8.8            & 123.4M & \textbf{0.721}     & 0.885             & \textbf{0.749}    \\ 
\bottomrule
\end{tabular}
\end{center}
\end{table}
\setlength{\tabcolsep}{2pt}

\begin{sidewaysfigure}

\includegraphics[width=\linewidth]{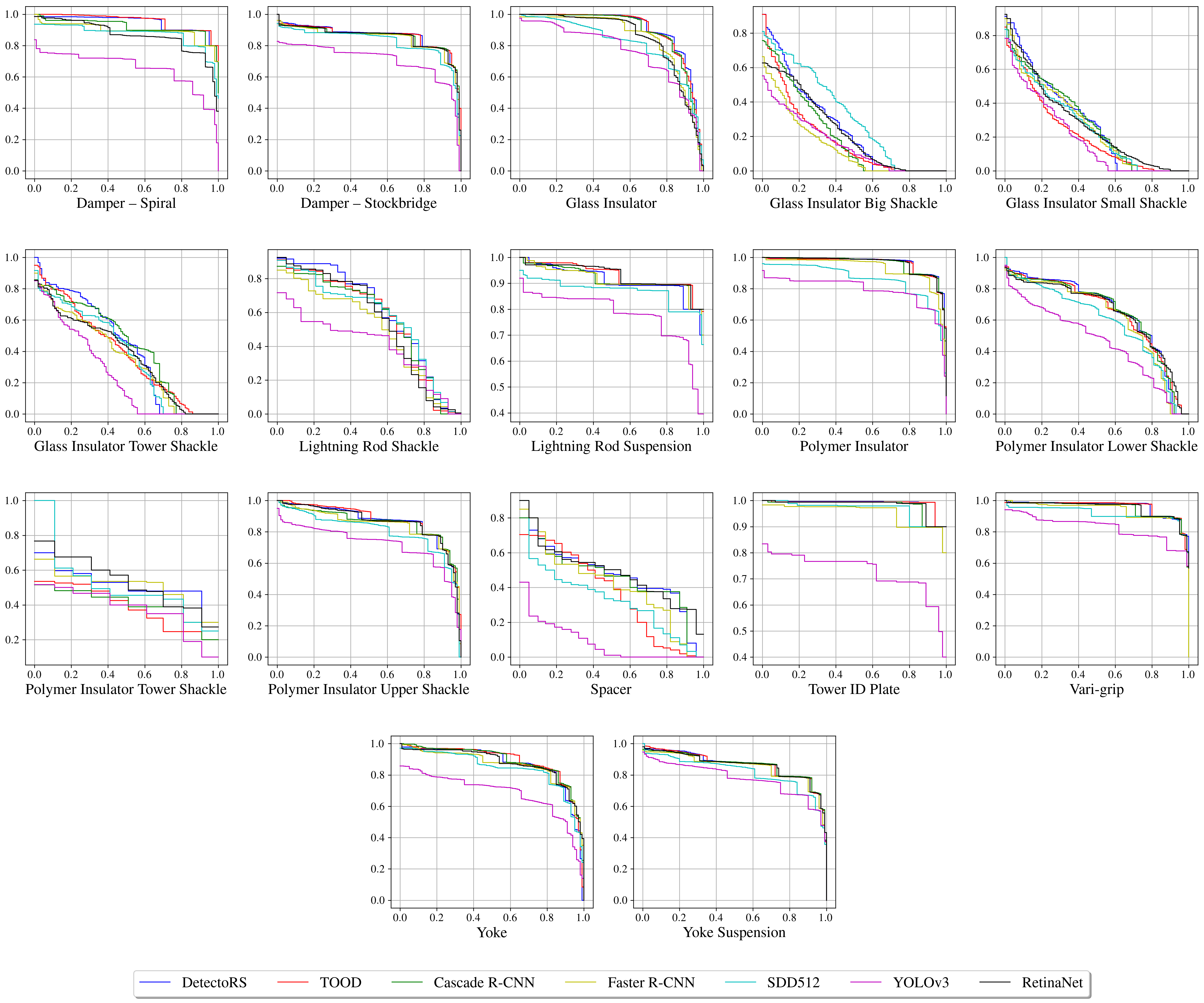} 
\caption{Precision-Recall curves for each InsPLAD-det asset and all evaluated object detection methods}
\label{fig:prcdetect}

\end{sidewaysfigure}

Figure~\ref{fig:prcdetect} shows the corresponding Precision-Recall curves for each asset and object detection method. The Dampers, Insulators, Tower ID Plate, Vari-grip, Lighting Rod Suspension, Yoke, and Yoke Suspension presented the best performance by the object detectors. At the same time, most types of Shackles and the Spacer had lower results. That can be explained mainly by the difference in the number of samples of each object. Almost all the objects the detectors perform best have the highest amount of samples in the dataset, which is expected since most Deep Learning methods are data-hungry. However, the Polymer Insulator Lower Shackle does not follow this pattern. It has a high amount of samples, but the object detectors do not present a good performance compared to other objects with many samples. That may be because different Polymer Insulator Lower Shackles instances have slight variations that do not justify separating them into sub-classes.

\begin{figure}

\begin{subfigure}[b]{\textwidth}
\includegraphics[width=\linewidth]{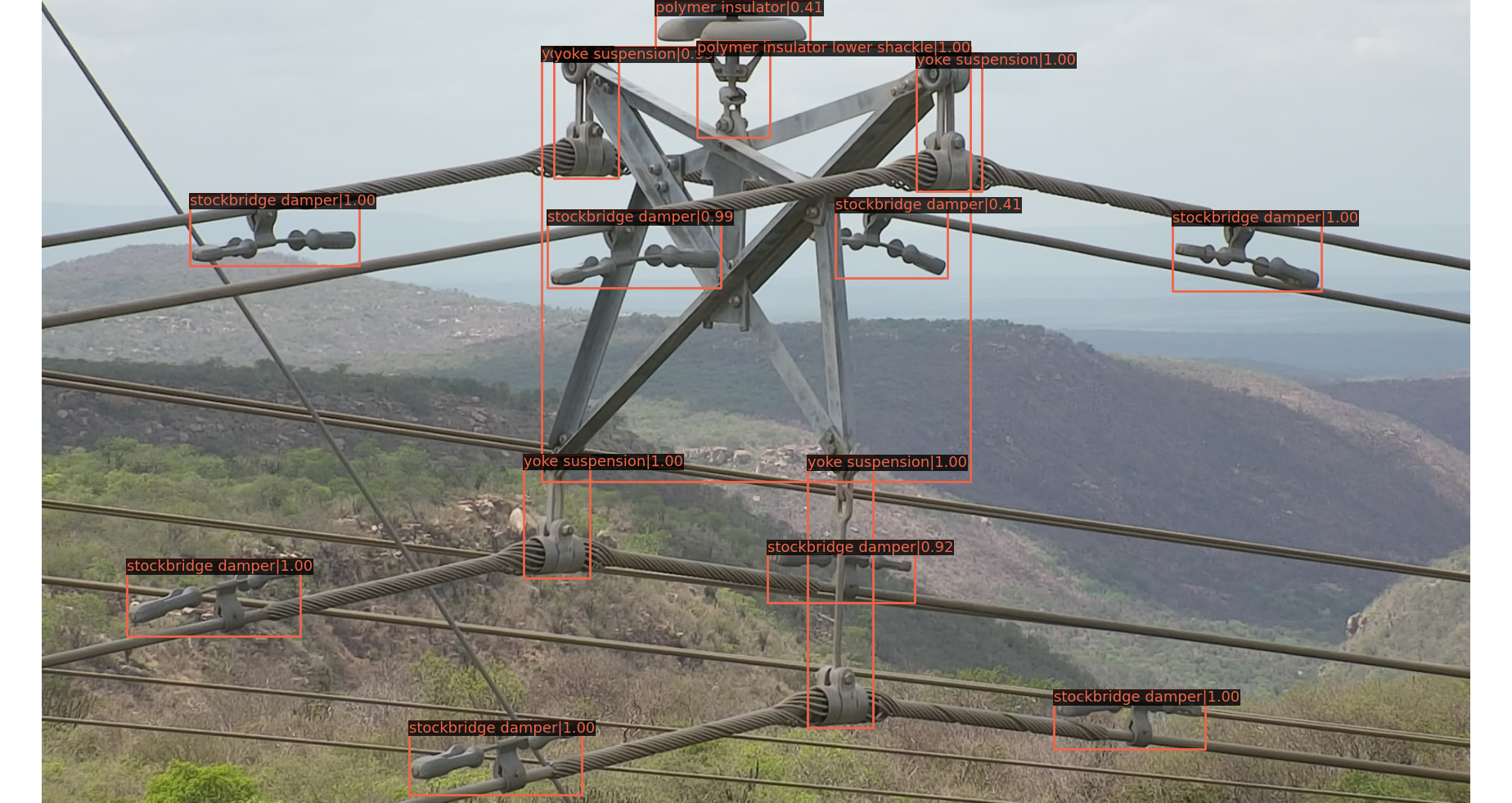} 
\caption{Successful case}
\label{fig:suc}
\end{subfigure}
\begin{subfigure}[b]{\textwidth}
\includegraphics[width=\linewidth]{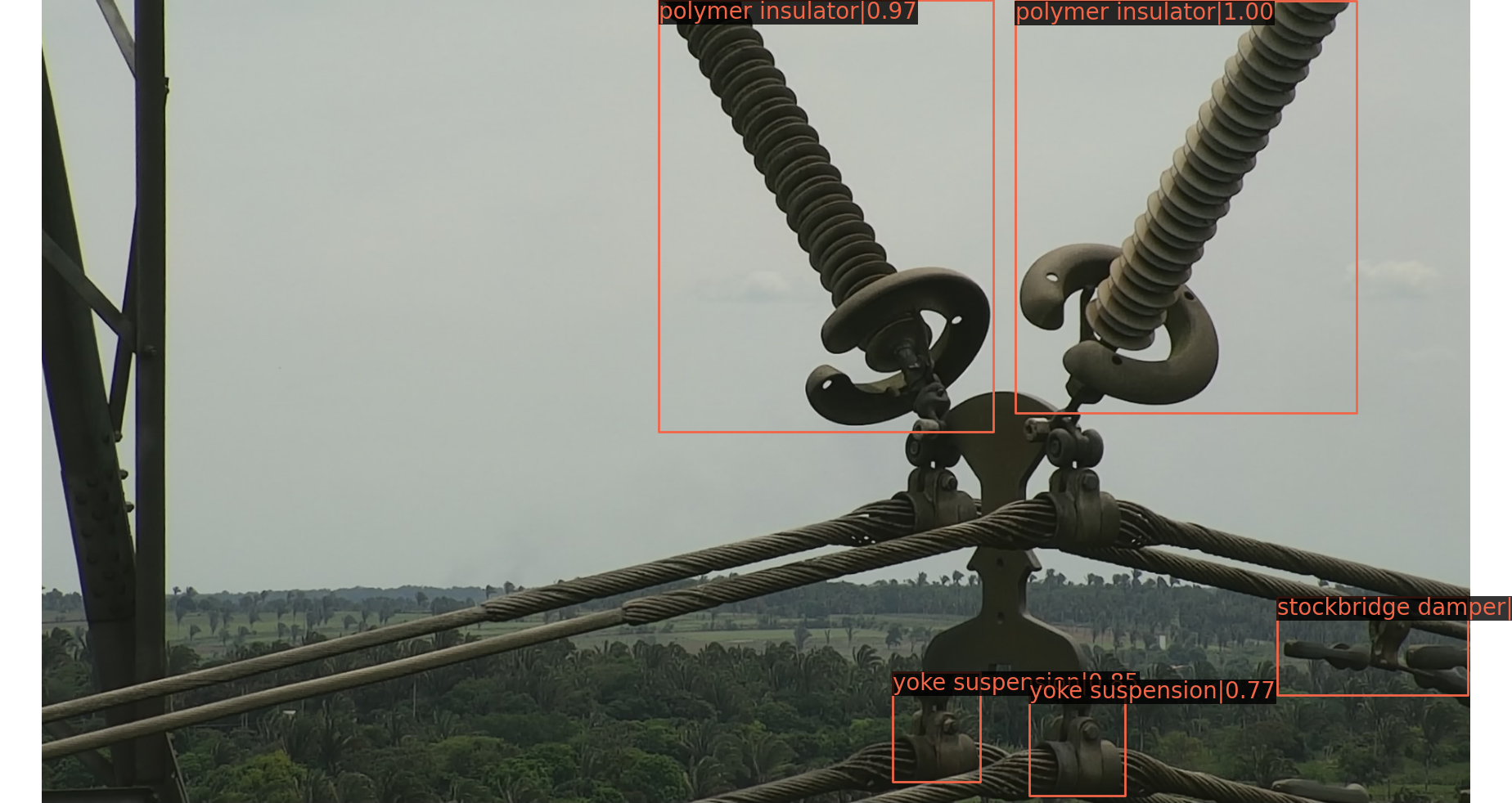} 
\caption{Failure case}
\label{fig:fai}
\end{subfigure}

\caption{DetectoRS results for two test set samples. \ref{fig:suc} is a successful case with multiple assets from five classes: Stockbridge Damper, Yoke Suspension, Yoke, Polymer Insulator, and Polymer Insulator Lower Shackle. In \ref{fig:fai}, two unoccluded objects were not detected: the Polymer Insulator Lower Shackle below the left Polymer Insulator and the Yoke Suspension below the right Polymer Insulator}
\label{fig:od_results}
\end{figure}

Figure~\ref{fig:od_results} shows the detection results of DetectoRS, which is the best-performing object detector in terms of Box AP. The first sample is a case where the detector is successful, and the other is an example of a failure. It can be noticed in the first sample that it performs well in a sample with multiple objects of different classes, including partially occluded objects, e.g., the Stockbridge Damper near the centre of the image. On the other hand, the fail case shows four objects that were not detected, two of them completely unoccluded: the Polymer Insulator Lower Shackle below the left Polymer Insulator and the Yoke Suspension below the Polymer Insulator to the right. That might be due to lighting issues since the objects are darker than usual or some minor variations in their structure compared to other instances.

\subsection{InsPLAD-fault}

\subsubsection{Supervised Fault Classification}
\label{sec:fault}

The second challenge presented by our dataset is an image classification task: given a detected power line asset, classify it regarding its apparent conditions. The asset conditions are varied, e.g., rust, missing cap, and being covered by a bird's nest; their extension depends on the asset category. We found those conditions in five of the seventeen classes detected in InsPLAD. For this purpose, we train popular and state-of-the-art supervised image classification methods, namely EfficientNet \cite{tan2019efficientnet}, ResNet \cite{he2016deep}, ResNeXt \cite{xie2017aggregated}, MLP Mixer \cite{tolstikhin2021mlp} and Swin Transformer \cite{liu2021swin}. The first three are CNN-based methods, while the other two use different approaches. Swin Transformer is based on Vision Transformers, which have recently become popular, while MLP-Mixer is a new computer vision method, purely based on Multi-layer Perceptrons, achieving competitive results on image classification benchmarks \cite{tolstikhin2021mlp}.

\paragraph{Setup}

Here, traditional data augmentation was used, increasing the training data by ten times, as seen in Subsection~\ref{sec:properties}. Finally, the following operations were randomly applied using the Albumentations library \cite{buslaev2020albumentations}: brightness and contrast alteration, histogram equalization, colour alteration by PCA, blurring, sharpening, horizontal mirroring, and rotation.

All object detection methods were evaluated using their implementations from the MMClassification toolbox, also by OpenMMLab, and also publicly available on GitHub \cite{MMClassification_Contributors_OpenMMLab_s_Image_Classification_2020}. They were initialised from publicly available pre-trained weights provided by the MMClassification toolbox and fine-tuned until loss stabilization. The methods were trained in their standard configurations of parameters and hyperparameters for EfficientNet-B2, ResNeXt-101 32x8d, ResNet-101 8xb32, MLP Mixer-Base 64xb64, and Swin Transformer-Tiny 16xb64, as found in MMClassification. For these default MMClassification settings, the models were trained in 100 epochs, input images with a resolution of 224x224, a batch size of 32, and a Learning Rate of 0.1, except for Swin Transformer, which required 300 training epochs, batch size of 64 and Learning Rate of 0.001.

Finally, the best-evaluated checkpoints were selected for further comparison by the end of training. Balanced accuracy is used for comparison since some test sets are imbalanced \cite{brodersen2010balanced}. Additionally, the size of weights and throughput were also registered.

\paragraph{Results and Discussion}

Table~\ref{table:faultcls} shows the results for each image classification method. EfficientNet achieves the best performance with a 0.954 average balanced accuracy. Swin Transformer presents a superior balanced accuracy in three out of five asset categories, with MLP-Mixer obtaining the highest for the Glass Insulator class and EfficientNet for the Vari-grip. Swin Transformer is the state-of-the-art method here, however, its average balanced accuracy is not able to surpass two other methods. On the other hand, MLP-Mixer is not based on CNNs or Vision Transformers and achieves performance comparable to the state-of-the-art. 

Regarding inference speed, all models show high potential to be used in real time. However, EfficientNet produces the lightest model approximately five times compared with the second most lightweight. In a real-time inspection scenario with several power line assets, this lightweight property may be an essential feature to save resources. Another advantage of EfficientNet in this scenario is its performance. In our comparison, EfficientNet achieves a 0.954 overall balanced accuracy, being the best overall by a small margin. 


\setlength{\tabcolsep}{2pt}
\begin{table}
\begin{center}
\caption{Fault classification balanced accuracy comparison between image classification methods for each asset class. Model size and throughput are also shown. The best results are highlighted in bold}
\label{table:faultcls}
\begin{tabular}{lccccc}
\toprule
                                & \makecell{ResNet \\ 
                                } & \makecell{ResNeXt \\ 
                                } & \makecell{EfficientNet \\ 
                                } & \makecell{MLP-Mixer \\ 
                                } & \makecell{Swin \\Transformer  \\
                                } \\ 
\midrule
Reference & CVPR16 & CVPR17 & ICML19 & NeurIPS21 & ICCV21 \\
Weights size (MB)               & \multicolumn{1}{c}{357}                & \multicolumn{1}{c}{712}                 & \textbf{74}                       & \multicolumn{1}{c}{719}                  & \multicolumn{1}{c}{341}                        \\
Inferences s$^{-1}$                    & \multicolumn{1}{c}{587}                & \multicolumn{1}{c}{347}                 & \multicolumn{1}{c}{544}                      & \textbf{589}                  & \multicolumn{1}{c}{587}                        \\
N° of parameters                    &         44.5M        &  88.8M                &   \textbf{9.1M}                    &         59.9M          &            28.3M             \\
\midrule
\makecell[l]{Glass Insulator}                 & 0.833                                  & 0.811                                   & 0.866                                        & \textbf{0.916}                                    & 0.883                                          \\
\makecell[l]{Lightning Rod Susp.}        & 0.989                                  & 0.926                                   & 0.996                                        & 0.987                                    & \textbf{0.998}  \\ 
\makecell[l]{Pol. Ins. Upper Shackle} & 0.832                                  & 0.876                                   & \textbf{0.985}                                        & 0.955                                    & \textbf{0.985} \\
Vari-grip                       & 0.690 & 0.500  & \textbf{0.953} & 0.907 & 0.889 \\
\makecell[l]{Yoke Suspension}                 & 0.861                                  & 0.832                                  & 0.970                                        & 0.992                                    & \textbf{0.999} \\
\midrule
Average                         & 0.841                                  & 0.789                                   & \textbf{0.954}                                        & 0.951                                    & 0.951                                
\\
\bottomrule
\end{tabular}
\end{center}
\end{table}
\setlength{\tabcolsep}{2pt}



\begin{figure}
\centering
\includegraphics[width=.5\linewidth]{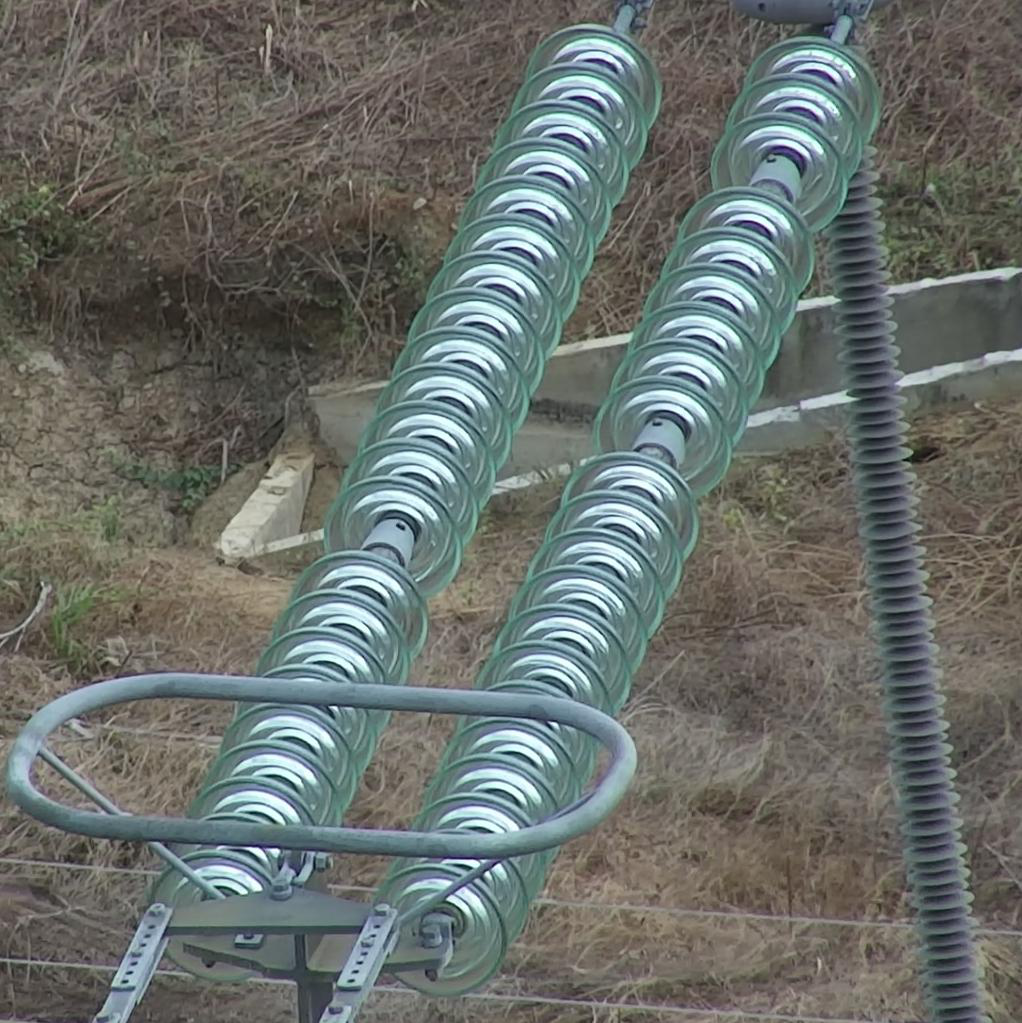} 
\caption{Failure case of EfficientNet for a test sample of InsPLAD-fault. It was incorrectly classified as a defect-free Glass Insulator}
\label{fig:effifail}

\end{figure}

Figure~\ref{fig:effifail} shows an example of a misclassified Glass Insulator by the EfficientNet method, which is the best-performing supervised fault classifier, on average. However, the method incorrectly classified it as a normal Glass Insulator, not a faulty one. That may be due to the Glass Insulator perspective, where the missing cap gaps are less evident than in a frontal view. However, the three missing caps are still recognizable by a specialist and even by the average human observer.

\subsubsection{Unsupervised Anomaly Detection} 
\label{sec:anomaly}

The third and final challenge covered by InsPLAD is an anomaly detection task: detect on an image level whether a (cropped) power line asset is normal or anomalous. In this task, the model is only trained with normal objects and tested with normal and anomalous objects. Here, all the asset conditions are treated as anomalies. The same five asset categories in the subsection above are selected since anomalous samples are required for achieving quantitative results when testing. For this task, 
we selected and fine-tuned four anomaly detection methods for our analysis. We chose the Autorencoder $L_2$ \cite{bergman2019improving}, because it was the best-performing method used in MVTec AD paper \cite{bergmann2019mvtec} to benchmark their proposed anomaly detection dataset. Secondly, OGNet \cite{zaheer2020old} was selected to check if one-class classifiers could yield better results given the uncontrolled environment property provided by InsPLAD. Finally, DifferNet \cite{rudolph2021same} is a recent method that presents good performance on the MVTec AD dataset benchmark, and CS-Flow \cite{rudolph2022fully} is a state-of-the-art anomaly detection method developed from DifferNet.

The first one is based on Convolutional Autoencoders \cite{goodfellow2016deep} and reconstructs patches of $128\times128$ employing a per-pixel $L_2$ distance Loss with a latent space dimension of 100. On the other hand, OGNet is based on GANs. The discriminator task is to distinguish between good and bad quality reconstructions yielded by a new and an old state of the generator, respectively. That allows the discriminator to learn the small nuances of an anomaly, outputting what is considered an anomaly score. Finally, the CS-Flow and DifferNet methods are based on CNNs and normalizing flows. They use the CNN feature extraction to obtain an anomaly score through normalizing flows. CS-Flow and DifferNet are specific defect detection methods tested on a state-of-the-art anomaly detection dataset \cite{bergmann2019mvtec}.

\paragraph{Setup}


Both feature-extraction CNNs in CS-Flow and DifferNet, EfficientNet-B5, and AlexNet, respectively, are initialised using weights pre-trained with ImageNet. The codes used in this experiment were obtained in their official repositories by their authors. Table~\ref{tab:anomalysetup} shows the experimental configurations for each method. As in the previous tasks, the best-evaluated checkpoints during training were selected for the baselines.

\begin{table}
\caption{Experimental configurations of the anomaly detection methods}
\centering
\label{tab:anomalysetup}
\begin{tabular}{@{}lcccc@{}}
\toprule
Method    & Input size & Epochs & Batch size & Learning rate     \\ \midrule
Autoencoder & 128x128    & 200    & 16    & 0.0002 \\
OGNet     & 45x45      & 200    & 16    & 0.001  \\
DifferNet & 448x448    & 192    & 24    & 0.0002 \\
CS-Flow   & 768x768    & 240    & 14    & 0.0002 \\ \bottomrule
\end{tabular}
\end{table}

The metric adopted for the baselines is the \ac{auroc} curve, which is the primary metric used by the community. The ROC curve is plotted with TPR against the FPR by varying the binary classification threshold for anomalous and normal using the anomaly score. As the curve is constructed from the TPR and FPR rates, it is robust to unbalanced datasets of anomalous and normal samples, which commonly occurs in anomaly detection application scenarios. Therefore, when there is a threshold value capable of separating true positives from true negatives, the curve nears the upper left corner, resulting in the area under the curve close to $1.0$, which means a good result. The curve nears a diagonal straight line in the opposite case, indicating a random classifier. In addition, the size of weights and throughput were also registered.

\paragraph{Results and Discussion}

\begin{figure*}

\includegraphics[width=\linewidth]{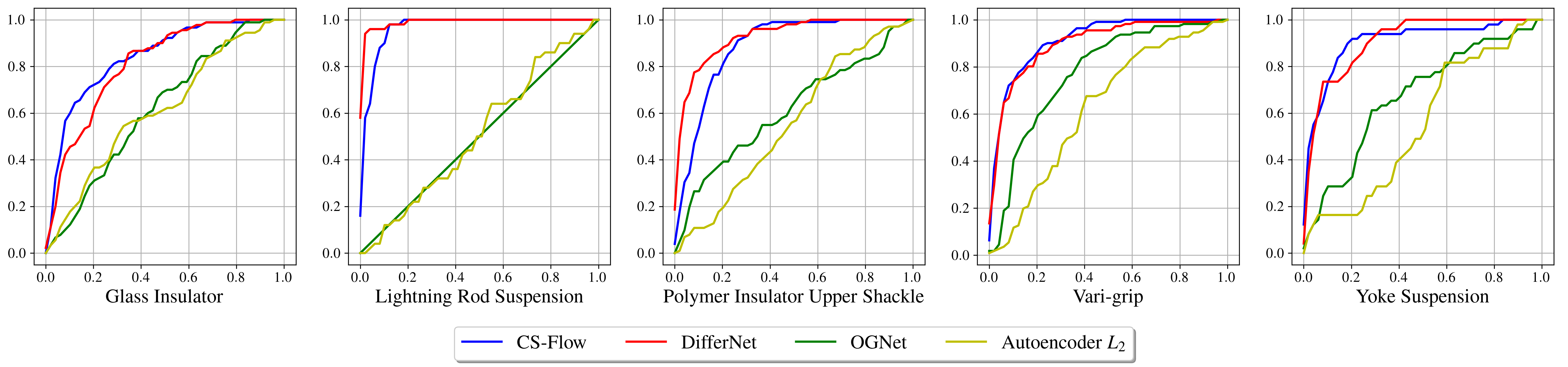} 
\caption{ROC curves, which plot True Positive Rate against False Positive Rate for each InsPLAD-fault category and all evaluated anomaly detection methods}
\label{fig:roc}

\end{figure*}

Table~\ref{tab:anomaly} summarises the results obtained for the anomaly detection methods in InsPLAD-anomaly. DifferNet achieves the best average AUROC result with 0.905, but CS-Flow obtains the best results in three out of five asset categories and comes in a close second with a 0.903 average AUROC. Due to that, there is no clear advantage to a single method. However, the approach using CNN feature extraction with normalizing flows adopted by both presents a superiority over the others. For the OGNet network, the average obtained among the five objects was 0.635. Based on adversarial training, the network obtained a better result than Autoencoder $L_2$, even with fewer parameters and used images with just 45 pixels on each side. Resizing, characteristic of the network pipeline, reduces the image, so the network needs fewer parameters to process it but loses information in the image. The loss of information makes training challenging to learn some object features, as occurred for the Lighting Rod Suspension, where the network training could not converge and classified every test sample as anomalous. Although Autoencoder $L_2$ is the lightest model, there is a more than 30 percentage points gap between it and the best-performing methods. For each object and anomaly detection method, the corresponding ROC curves are given in Figure~\ref{fig:roc}.

In terms of inference throughput, all methods achieved satisfactory results. Although DifferNet is the largest model, it is the best-performing one, making 70 inferences per second. It is a promising alternative in a real-time inspection scenario, considering other tasks must be executed in the inspection pipeline. 

\setlength{\tabcolsep}{3pt}
\begin{table}
\begin{center}
\caption{AUROC comparison between anomaly detection methods for each evaluated asset. The size of the models in megabytes and throughput are also presented. The best results are in bold}
\label{tab:anomaly}
\begin{tabular}{lcccc}
\toprule\noalign{\smallskip}
& \makecell{Autoencoder $L_2$ \\
} & \makecell{OGNet \\
} & \makecell{DifferNet \\
} & \makecell{CS-Flow \\
} \\
\noalign{\smallskip}
\hline
\noalign{\smallskip}
Reference                                                                                                      & VISAPP19                                    & CVPR20                                  & WACV21                                      & WACV22                                    \\
Weights size (MB)                                                                                             & \textbf{4}                                           & 52                                      & 933                                         & 602$\sim$727                              \\
Inferences s$^{-1}$                                                                                                   & \textbf{96}                        & 80                    & 70                                          & 39                                        \\
N° of parameters                                                                                                  & 28.8M                        & \textbf{12.9M}                    & 233.1M                                          & 148.2M                                        \\
\noalign{\smallskip}
\hline
\noalign{\smallskip}
Glass Insulator                                                                                                & 0.614                                       & 0.618                                   & 0.824                                       & \textbf{0.841}                            \\
\begin{tabular}[c]{@{}l@{}}Lightning Rod Suspension\end{tabular}                                                                                       & 0.512                                       & 0.500                                   & \textbf{0.991}                              & 0.966                                     \\
\begin{tabular}[c]{@{}l@{}}Pol. Insulator Upper Shackle\end{tabular} & 0.555                                       & 0.612                                   & \textbf{0.901}                              & 0.884                                     \\
Vari-grip                                                                                                      & 0.628                                       & 0.775                                   & 0.906                                       & \textbf{0.915}                            \\
\begin{tabular}[c]{@{}l@{}}Yoke Suspension\end{tabular}                                & 0.551                                       & 0.672                                   & 0.905                                       & \textbf{0.907}                            \\
\noalign{\smallskip}
\hline
\noalign{\smallskip}
Average                                                                                                        & 0.572                                      & 0.635                                   & \textbf{0.905}                              & 0.903                                    
\\\bottomrule
\end{tabular}
\end{center}
\end{table}
\setlength{\tabcolsep}{3pt}

\subsection{InsPLAD Strengths and Limitations}

The InsPLAD is proposed in a context where multiple works acknowledge a lack of public benchmarks for power line inspection. As presented in the Introduction, each challenge proposed by InsPLAD aims to fill the gaps mentioned in Table \ref{tab:contributions}. Firstly, InsPLAD-det presents many categories of power line assets under various external conditions, orientations, and subtypes, with more than 28k samples. That is much higher than the current public power line asset data. However, the amount of samples for each class varies significantly, producing an unbalanced dataset. To help with the imbalance between classes, data augmentation techniques, as also done in MVTec AD, help reduce the impact caused by the frequency diversity of objects in transmission line towers. An example of performing data augmentation in this work is applying rotation and mirroring randomly to generate new images of specific classes.

InsPLAD-fault and InsPLAD-anomaly share similar advantages and drawbacks. Public data regarding faulty power line assets is even rarer than datasets for object detection or instance segmentation of power line assets. Both InsPLAD datasets present different faults/anomalies for different assets, which is unprecedented. They contain six defects: four are corrosion, one is a broken component, and one is the presence of a bird's nest. However, these defects are spread over five assets; four have just one type of fault, and one has two types: corrosion and bird's nest presence.
That arises from the inherent aspect that finding anomalies/defects in real-world operating power lines is challenging, as they receive predictive maintenance and the problems are scarce and quickly corrected. Therefore, it would be ideal to have more than one type of defect per category, as seen in other anomaly detection datasets, for instance, MVTec AD, which focuses on objects in controlled environments and has on average five per category \cite{bergmann2019mvtec}. As an anomaly is defined as something that deviates from the standard, containing any deviation type, it is important to have variations with diversity to ensure that the proposed models learned the object's standard and not how to identify a specific problem. With defects similar to each other or just one type of defect for the asset, it allows a model to learn to recognize that defect and not the normal pattern expected from the asset. As shown by \cite{campos2016evaluation} for Outlier detection, a similar problem, datasets with diversity in their outlier examples allow a new method to demonstrate its superiority over existing methods by consistently detecting the outliers in these datasets.

Finally, both InsPLAD-det and InsPLAD-fault share similar challenges with other computer vision application areas, such as agriculture with the detection of plant diseases \cite{singh2020plantdoc}, in transport with the detection of pedestrians and vehicles \cite{caesar2020nuscenes}, in sport with the analysis of player performance \cite{wu2022sports}, among others. All these areas and the Powerline inspection have challenges in common, such as small and big objects, multiple objects per image, uncontrolled background, environment and point-of-views, and objects with multiple instances.    

\section{Conclusions}

This work proposes InsPLAD, a dataset for power line asset inspection made from UAV images from real-world inspections. InsPLAD is organised to meet different steps on a standard power line inspection: object detection for the detection of power line components in UAV images, image classification for identifying defects in the detected components, and unsupervised anomaly detection, which has the same objective as the latter but treats the defects as anomalies and no defective data is used for training. InsPLAD is the largest public power line-related dataset regarding power line components, number of images, annotations, defects, and vision tasks. A benchmark with state-of-the-art and popular methods for object detection, image classification, and anomaly detection is also provided to serve as a baseline. It shows room for improvement in all three steps. We hope InsPLAD sparks future research on visual inspection in the power line area and other areas with similar challenges.

InsPLAD opens lots of possibilities for future research. More specifically, new methods for unsupervised anomaly detection in the wild seem quite attractive since defective component data is scarce, and new datasets alone will not be able to cover all possible defect types due to the uncontrolled environment nature of the problem. Besides that, InsPLAD shows room for improvement in it. Another problem is the large size of the best-performing anomaly detection methods, which heavily hinders their deployment into production, even more so considering that a different model is used for every component type. Finally, new datasets for power line inspection are also welcome, especially ones with data on multiple defective components.

\section*{Acknowledgement(s)}

The authors acknowledge the financial support of STN - Sistema de Transmissão Nordeste S.A. through the ANEEL R\&D Program for the development of the research project entitled: “PD-04825-0006/2019: Inspeção com Drones por Meio do Acoplamento Eletrostático para Carregamento de Baterias em Voo e Uso de Aprendizagem de Máquina para Classificação Automática de Defeitos”. 

\section*{Disclosure of interest}

 The authors report no conflict of interest.

\section*{Funding}

This work was supported by STN - Sistema de Transmissão Nordeste S.A. through the ANEEL R\&D Program for the development of the research project under grant code PD-04825-0006/2019; Coordenação de Aperfeiçoamento de Pessoal de Nível Superior - Brasil (CAPES) under Finance Code 001; and Conselho Nacional de Desenvolvimento Científico e Tecnológico (CNPq).

\section*{Data availability statement}

The data that support the findings of this study is openly available in Mendeley Data at \url{https://data.mendeley.com/preview/5n3fjgvfyz?a=f68efe15-61d3-4a74-bc8a-d009e3cd3f95}.

\bibliographystyle{chicago}
\bibliography{interactcadsample}

\end{document}